# Emerge-Sort: Converging to Ordered Sequences by Simple Local Operators

March 10, 2009

Dimitris KALLES[1], Alexis KAPORIS[2]

**ABSTRACT**

In this paper we examine sorting on the assumption that we do not know in advance which way to sort a sequence of numbers and we set at work simple local comparison and swap operators whose repeating application ends up in sorted sequences. These are the basic elements of Emerge-Sort, our approach to self-organizing sorting, which we then validate experimentally across a range of samples. Observing an $O(n^2)$ run-time behaviour, we note that the $n/\log n$ delay coefficient that differentiates Emerge-Sort from the classical comparison based algorithms is an instantiation of *the price of anarchy* we pay for not imposing a sorting order and for letting that order emerge through the local interactions.

## 1 Introduction

Sorting has been one of the first areas of computer science that witnessed the design of very efficient algorithms to stand the test of time. While these algorithms (quick-sort, merge-sort, heap-sort) are good at sorting with relatively few constraining assumptions, if any, research focus gradually shifted to designing algorithms that can harness special cases in the input to accommodate some practical settings (for example, input that is partially ordered, as described by Manilla (1985), or drawn from a constrained distribution, as is the case in bucket-sort and spread-sort). Moreover, researchers have also studied extremely inefficient sorting algorithms to see whether inefficiencies can be mitigated by very simple changes in the algorithms. In this paper we examine sorting from the point of view of *relaxing the assumption that we know which way to sort* and attempting to see *what seems to be the minimum design of local operators so that a repetitive application of such operators ends up in sorted sequences*.

---

[1] Hellenic Open University and Open University of Cyprus, email: dkalles@acm.org

[2] Hellenic Open University and Research Academic Computer Technology Institute (Rio, Greece), email: kaporisa@gmail.com

Our interest in studying local operators is two-fold.

First, we are aware of the huge research effort that is been directed at studying various computational games from the point of view of reaching a Nash equilibrium; therein, we usually favour algorithms that make little use of global knowledge and where agents act competitively to each other yet manage to converge to a state that satisfies them all (Koutsoupias & Papadimitriou, 1999; Kalles *et al.*, 2008). In such settings, we are also usually looking to find out what expense such anarchy incurs when seen from an optimisation point of view; in other words, if we could centrally design what each agent will do, we are interested in knowing how much effort we would save. So, we are asking: what is the cost of not knowing which way we need to sort and how can we induce our sequences to self-sort?

Secondly, the swarm intelligence approach to problem-solving has recently gathered significant momentum since many traditional problems (graph partitioning, clustering) have been recast in terms of swarm behaviour. Swarm intelligence is at the junction of randomized behaviour and local operations, without a conventional pay-off motive as in the games domain, and has been demonstrated to solve problems such as task scheduling (Bonabeau *et al.*, 1998).

As a matter of fact, emergent sorting is a behaviour that has been documented in insect societies and modelled via swarm intelligence principles (Bonabeau *et al.*, 1999), albeit in an unconventional way; therein, sorting takes place in 2-D and consists of forming concentric circles where items of similar size are at roughly the same distance from the centre. 2-D sorting has received relatively scant attention since it is usually seen as a (difficult) case of clustering, another classic problem that has been also recently addressed with swarm intelligence (Handl and Meyer, 2008). It is also interesting to note that swarm intelligence is now being tied to autonomic computing research, mainly through the observation of autonomic computing principles in biology inspired systems and the transfer of the relevant concepts to problems in distributed computing (Babaoglu *et al.*, 2006).

Seen from a more conventional computer science perspective, swarm intelligence indeed draws on several aspects of distributed computing. Thus it is not surprising, that distributed sorting algorithms have been designed and progressively refined to address related problems of increasing relaxation of assumptions. We thus witness Loui (1984) first framing the problem of distributed sorting across *n* nodes linked in a chain, which he based on a three-stage process of first electing a leader and then, using a carefully crafted encoding of the numbers to be sorted, a second stage of inserting values in the right place and letting all nodes know about that encoding, and a third stage of selectively deleting values from all nodes, so that each node is left with the proper value of the sorted order. Based on that idea, Gerstel and Zaks (1997) then extended the proofs of various properties of these distributed sorting algorithms to tree-structures of connected processors and, quite recently, Flochinni *et al.* (2004) tackled various difficult extensions of the basic problem, among which one also observes the un-oriented sorting which is the problem we are examining. Hofstee *et al.* (1990) did not use passes for their distributed sorting variant but, nevertheless, injected knowledge of the direction of sorting; such knowledge is explicitly built in even in the simplest parallel and sequential sorting algorithms like odd-even sort[3] and gnome sort.[4]

It is interesting to see that $O(n^2)$ seems to be the minimum one has to pay for such sorting though, in the distributed computing context, the cost of messaging is also taken into account and factored into slight variations of the underlying $O(n^2)$ complexity. But, we note that, while these approaches also tackle the problem of election, that very election is central to the sorting; in our case, we simply apply local operators which have no *understanding* of *process* in the sense of a sequence of actions. From that perspective, it is then cellular automata that seem to be also related to our approach to sorting (Gordillo and Luna, 1994).

The rest of this paper is structured as follows. In the next section we sketch out the basic principles that have lead to Emerge-Sort, our proposal for self-organising sorting, by describing showing how smaller or larger modifications in the local operators have influenced the capacity of the algorithm to finally deliver sorted sequences. We then have two sections on the experimental validation of Emerge-Sort, one that deals with the demonstration of the basic principles (or the lack thereof) and one that deals with behaviour on input that has some special characteristics, be it either partial order or controlled shuffling of elements in a sorted sequence. Finally, before concluding, we discuss the theoretical questions that are relevant for a future analysis of Emerge-Sort.

## 2 Developing Emerge-Sort

Let us take three distinct numbers and let us examine the possible ways in which these numbers may be permuted. Figure 1 depicts the possible relationships between the elements of any triple (*a*,*b*,*c*); we use that notation because it simplifies our reference to the local operators.

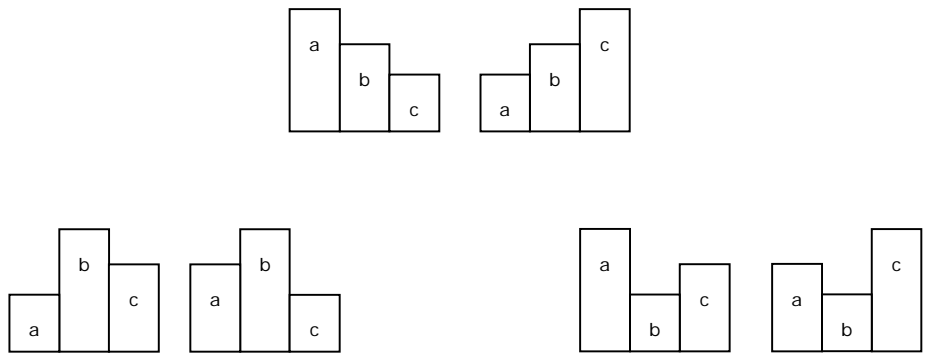

Figure 1: All possible permutations of three numbers.

---

[3] http://en.wikipedia.org/wiki/Odd-even_sort

[4] http://en.wikipedia.org/wiki/Gnome_sort

It is quite clear that when faced with a top row configuration, we are done provided we have no direction bias. It is also rather intuitive how to handle the lower row configurations: we swap those two elements that will result in a sorted triple. This is depicted in Figure 2.

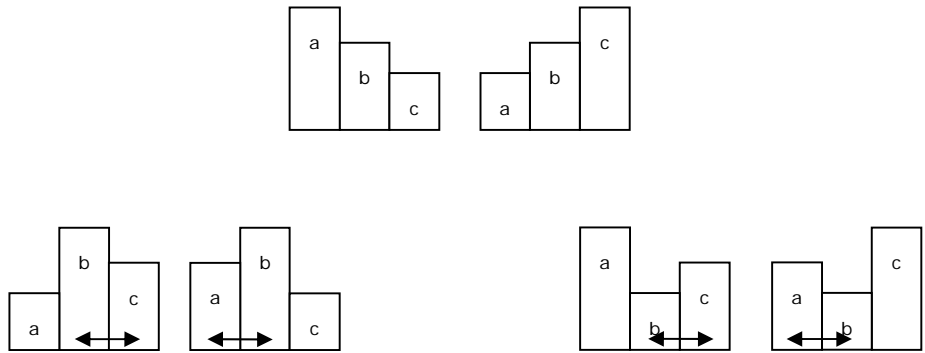

Figure 2: Required swaps to sort any three numbers.

It would be difficult to imagine a simpler operator on any three numbers. One could hope that this operator would be able to scale to arbitrarily selected triples of any unordered sequence of distinct numbers and that, eventually, would result in an ordered sequence.

It turns out that this is not the case (it is not difficult to devise a counter-example demonstrating eternal oscillation between two sequences); the next best attempt is to try to see how to extend those operators. We now set out the three key extensions we have considered and have delivered results:

- Allow numbers to demonstrate limited momentum (i.e. prefer to not change the direction you have been moving to).
- Examine triples in succession.
- Wrap the sequence when treating endpoints.

## 2.1 Injecting Momentum

The rationale behind momentum is rather obvious: after a triple has sorted itself, a subsequent neighbouring triple may trigger an oscillation to a previous state, which is a waste of resources. Figure 3 illustrates this point (the lightly shaded box represents a number that is not part of the examined triple).

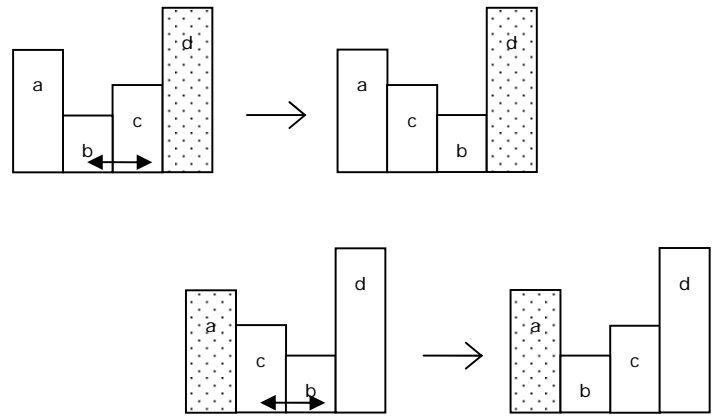

Figure 3: Demonstrating sorting oscillations.

While in the above example it does seem obvious that these four numbers could sort themselves anchored to *d*, oscillations in longer sequences can result in persistent peaks and valleys in the numbers. We need to find a way to postpone some swaps until we have evidence that we need them. This is the concept of *momentum*.

The simplest form of momentum is expressed in terms of a ternary flag, associated with each number. This flag can be either **L**(eft), **R**(ight) or **X** (Don't Care). All flags are initialized at *X*. The idea is that when we decide we need a swap, we eventually swap only if the momentum flags are compatible.

Momentum compatibility is easy to define: we have a problem when the flag of a number points to a different direction we want it to head to. So, in a swap, we may have compatibility problems due to both cells, but one would suffice to abort the swap. If we abort a swap, however, we change both numbers' flags to *X*.

If we eventually swap, we change both numbers' flags to *L* and *R* correspondingly. Moreover, to account for the new situation, we also change that pair's neighbours' flags to *X*.

We remind the reader that all the above apply after we have examined a triple and decided to plan a swap. The only case where they also apply before the planned swap and we detect momentum *before* deciding is when the largest number of the triple is in the middle (lower left part of Figure 1); in that case we plan a swap that is simply compatible with the momentum of the largest number.

## 2.2  Examining Sequence Triples in Succession

Initially it may sound rather counter-intuitive to consider that one might need a succession in a swarm-based approach to any problem; swarm and order are a relative contradiction of terms.

Yet, what actually happens is that (see Figure 4) we are attempting to capitalise on the momentum concept earlier described, but this time we do that at the sequence level. One can imagine a token being passed from triple to triple along a direction as opposed to being tossed in the air and whichever triple grabs it gets the right to sort itself.

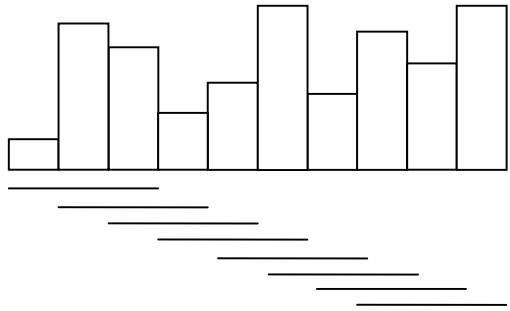

Figure 4: Demonstrating the succession of examined triples.

Note, again, that when we speak about triples, we are actually referring to triples of locations that hold numbers and not the numbers themselves. For example, with reference to Figure 4, note that a smallest number in the far left can eventually end up in the far right in one pass, participating in each examined triple, if the momentum flags so allow.

## 2.3 Sequence Wrapping

We have implemented a very simple wrapping that also happens to have an appealing visual metaphor (see Figure 5); essentially we are asking the numbers to sort themselves in a circle and indicate where the endpoints will be, without, however, indicating any preference for which extreme will be at which side of the separating line. To keep wrapping as simple as possible, we do NOT modify the momentum flag of a neighbour of a swapped number if we cross the separating line to reach that neighbour.

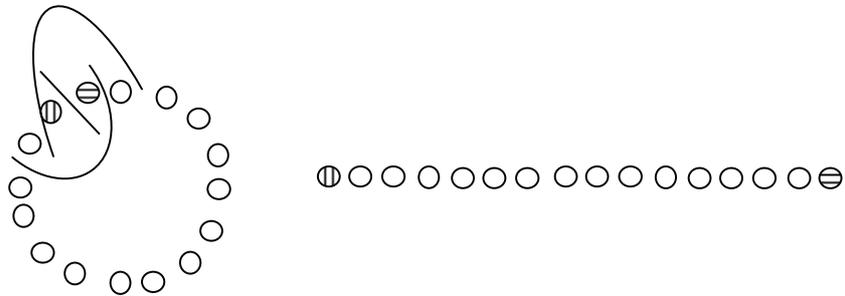

Figure 5: Demonstrating how endpoints wrap to form triples.

## 3 Experimental Validation of Emerge-Sort

We start be noting that we do not report experiments without momentum or without endpoint sequence wrapping; all these reached a point where numbers oscillated in peaks or valleys of the sequence out of which they were unable to escape. Moreover, we only experiment with sequences that have unique numbers, since this does not affect the generalization of our approach.[5]

Besides evaluating Emerge-Sort *per se*, we are interested in finding out what are the implications (or, consequences) of further diluting or strengthening the extensions set out above. The key reason is that we would like to observe whether, at a large scale, some modifications might deliver significant improvement at the expense of requiring some centralised control.

First, we note that in all the working variants of Emerge-Sort we observe an O($n^2$) run-time behaviour. While this is not to speculate that this could be analytically proved, we are, in the long term, interested to see whether the $n/\log n$ delay coefficient that apparently sets apart Emerge-Sort from quick-sort, heap-sort and the related genre of comparison-based algorithms, is *the price of anarchy* we pay for not imposing a sorting order.

## 3.1 Examining Sequence Triples in Succession – Does Direction Matter?

Figure 6 and Figure 7 confirm that it should not make a difference whether we traverse a sequence from left to right or vice versa. Both *LR* (Left-to-Right) and *RL* examinations are very close to each other throughout a large range of experiments[6], as regards both the average numbers of triples examined and the average number of moves per item (though there is a slight suggestion, which can be confirmed by looking at the actual numbers, that it takes increasingly longer to perform a move; note the slight sub-linear turn in the number of moves for large samples).

---

[5] There is a simple way to generate a data structure where no duplicate numbers are allowed; each number-to-be-sorted is associated with a count of its occurrences in the sequence. Whenever two adjacent numbers are equal, the sequence decreases by one number and the corresponding counters are added. While this needs some finer detail in the implementation of the operators, it does not affect the basic principles of momentum and wrapping and, therefore, we have not implemented it in the experiments that follow; instead, we generate sequences with unique numbers.

[6] Every data point is the average of ten experiments. This setting applies to all results reported in this paper.

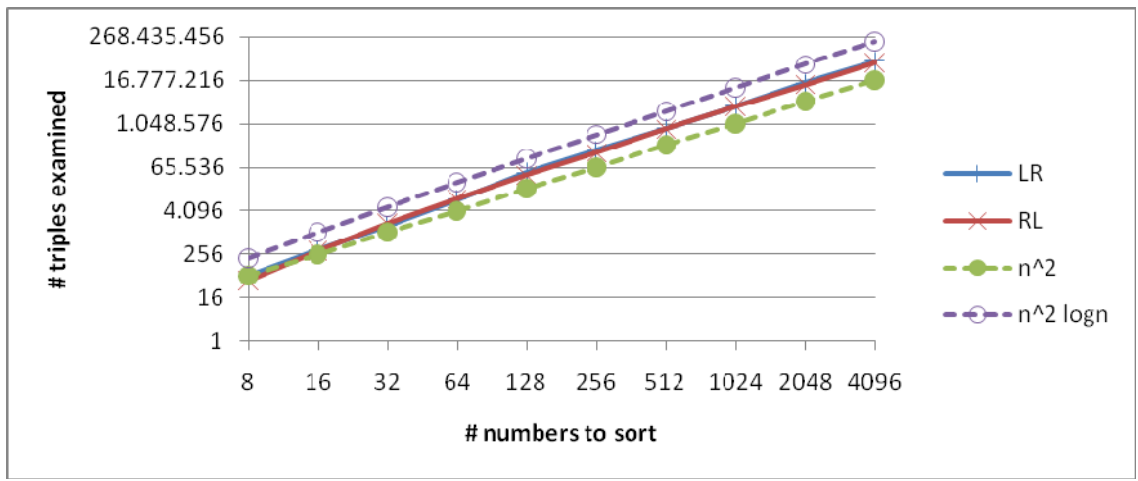

Figure 6: Direction does not affect the number of triples examined.

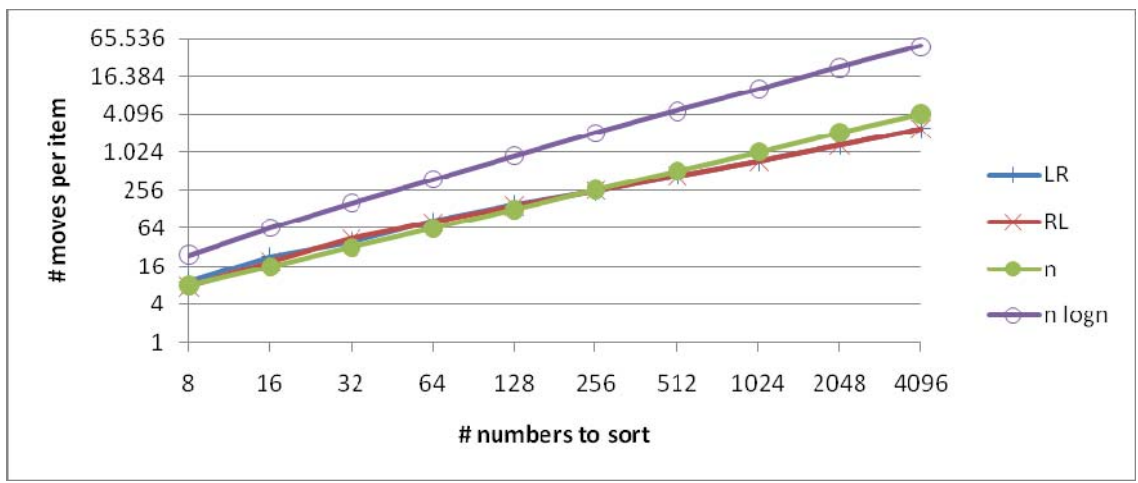

Figure 7: Direction does not affect the number of moves performed.

### 3.2 Why we Examine Sequence Triples in Succession

Let us now keep the *LR* variant described above and see how it is important to have a direction momentum at the sequence level. The first comparison has to do with having no sequence momentum at all. There are actually two ways to deal with this:

- Purely random: at any time point select a triple at random (*all-random*)
- Orderly random: examine all *n* triples in one batch, at random *within* that batch (*n-random*)

Essentially, the second ways ensures that during each batch, each triple will be examined once. We stress that both variants are full *LR* variants; with the exception of sequence momentum, they handle number momentum and sequence wrapping identically.

Observing Figure 8, we note that the *all-random* and *n-random* variants are close to O($n^4$) so we are pretty confident that the experimental agenda should not use these variants any longer.

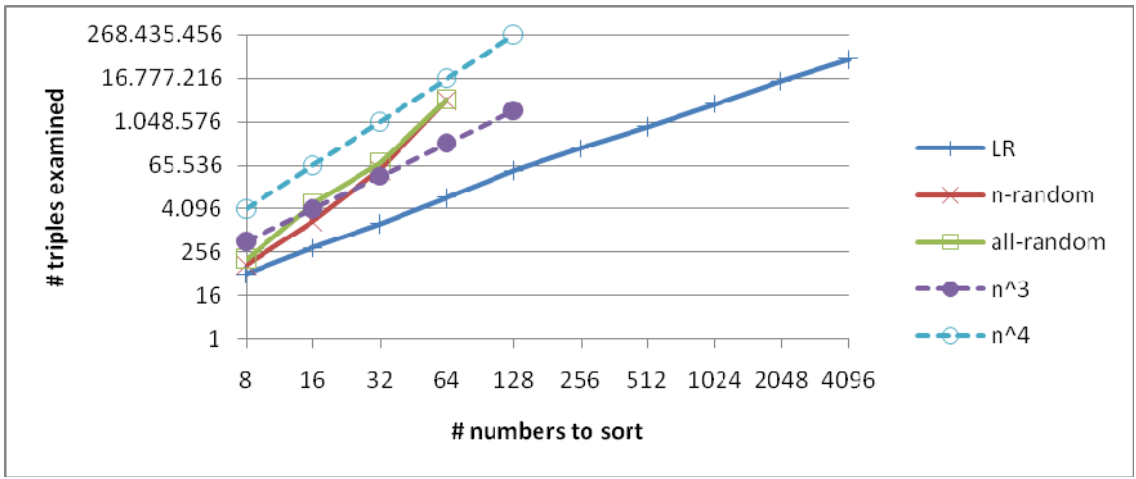

Figure 8: Demonstrating the need for sequence direction momentum.

However, there are two further variants we examine, and these have to do with whether the sequence momentum needs to be fixed:

- Random: whenever examining all triples in a succession, decide at random which way to traverse the sequence by tossing a fair binary coin – that's the *L_or_R(rnd)* variant
- Toggle: reverse the direction every time you need to decide on the direction - that's the *L_or_R(t)* variant

Observing Figure 9, we note that the *LR* baseline consistently outperforms the other two. For example, for $n$ = 4096 the value of *L_or_R(t)* is nearly twice that of *LR*, and this behaviour settles early on in the experimentation. In contrast, *L_or_R(rnd)* grows to being about five times more expensive than *LR* having started at being roughly equal.

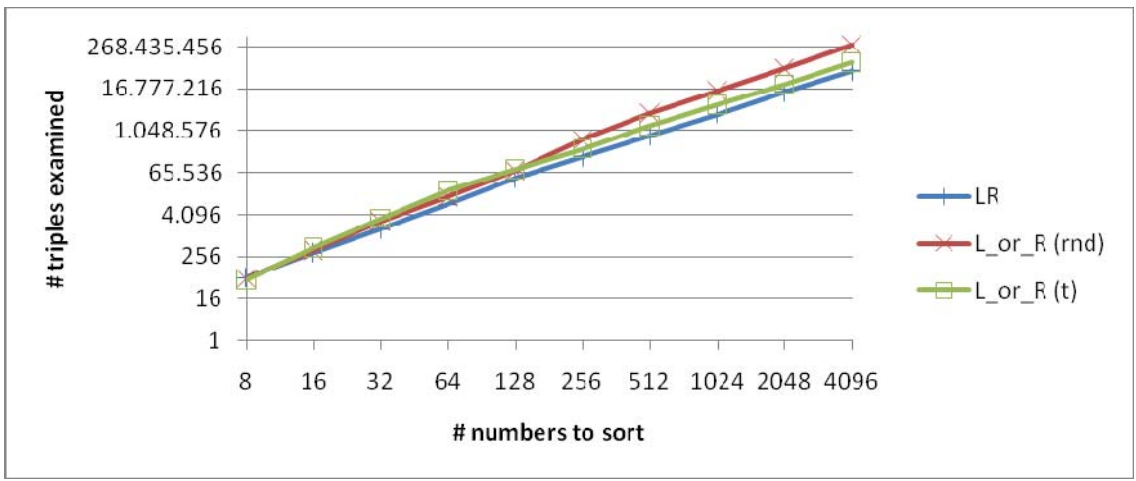

Figure 9: Demonstrating the need for consistent sequence direction momentum.

## 3.3 Demonstrating the Importance of Momentum in Examining Triples

We first show a simple experiment: what happens if we waive the rule that asks us to change a swap pair's neighbours' flags to *X*? This is the *LR(nnm)* variant (where *nnm* stands for no-neighbour-momentum).

Figure 10 shows the results and for the sake of comparison we have also included the *L_or_R(rnd)* variant, to show *LR(nnm)* is comparably inefficient in terms of triples examined.

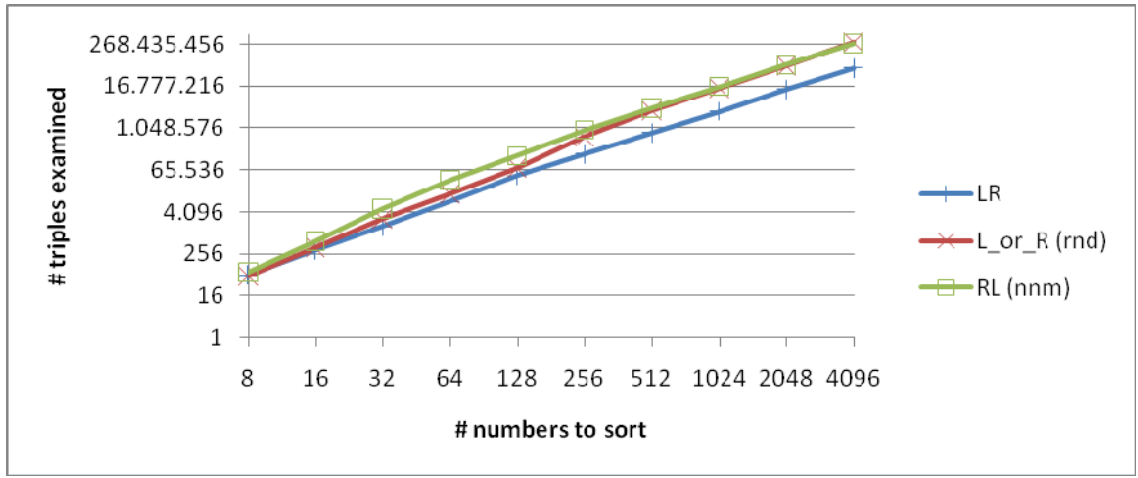

Figure 10: Not applying momentum to neighbours increases the number of triples examined.

However, if we also examine the number of moves required per item, to converge to a sorted sequence, we observe that not allowing the neighbours to update their momentum does not result in an as excessive number of moves to reach the final ordered sequence, as is the case for the *L_or_R(rnd)* variant.

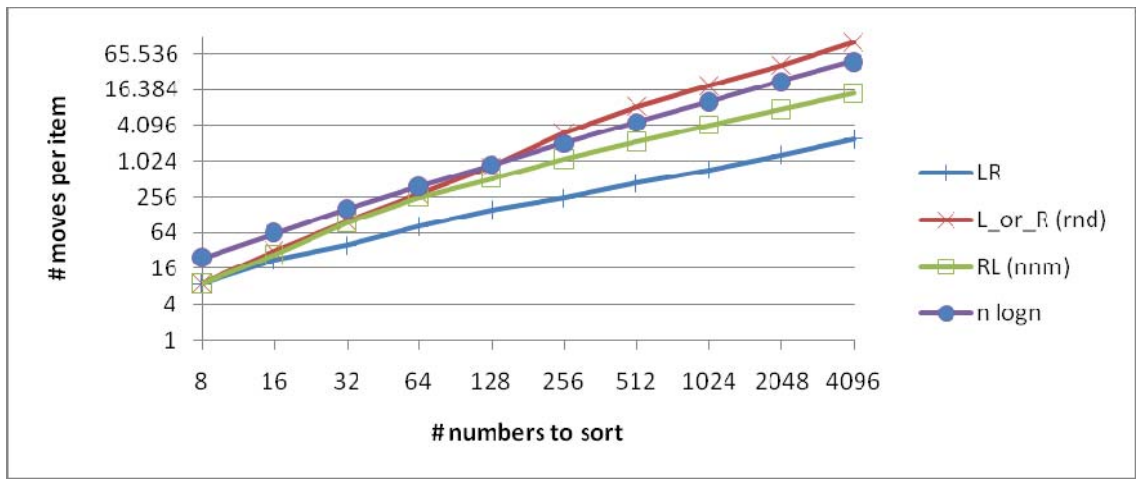

Figure 11: Not applying momentum to neighbours increases the number of moves required.

To shed more light into the behaviour of the *LR(nnm)* variant we computed the ratios for its performance compared to *LR*; we report the results in Table 1. Therein we note that *LR(nnm)* is relatively consistent with respect to *LR*, since we observe smooth analogies between how comparisons are associated with actual moves (swaps) on the way to ordering.

Table 1. Basic performance ratios for *LR(nnm)*/*LR* quantities

| *n* | # triples examined | # moves examined |
|---|---:|---:|
| **8** | 1,17 | 0,99 |
| **16** | 1,80 | 1,22 |
| **32** | 3,17 | 2,44 |
| **64** | 4,09 | 3,11 |
| **128** | 3,62 | 3,48 |
| **256** | 5,00 | 4,58 |
| **512** | 5,19 | 5,09 |
| **1024** | 5,15 | 5,72 |
| **2048** | 5,24 | 5,90 |
| **4096** | 5,27 | 6,10 |

Another question to ask is about what constitutes reasonable momentum at the number (triple) level. To experiment with this question we decided to simplify the momentum injection operator by only applying it to one number of a swap pair (neighbours still get an *X* flag). The two variants, therefore, are:

- Left: only affect the *new* left number of a swap pair – that's the *LR(mL)* variant
- Right: only affect the *new* right number of a swap pair – that's the *LR(mR)* variant

It is interesting to note that the *LR(mL)* variant has the counter-intuitive effect of affecting the momentum of the leftmost number, whereas it is in the right of the swap pair that the new action will take place. One should, then, expect that this should result in a delay. Figure 12 clearly confirms this expectation (we report the *RL(mR)* variant, instead).

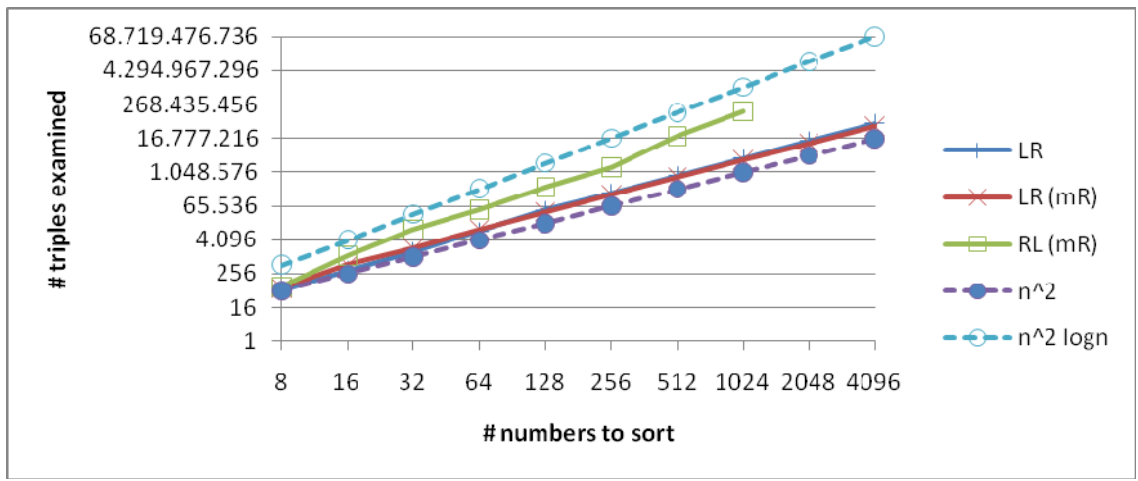

Figure 12: Constrained triple momentum must be compatible with sequence momentum.

On the other hand, the *LR(mR)* variant does comparably well (as a matter of fact it is slightly better than LR in terms of triples examined). However, it is not overall more efficient, since it asymptotically seems to generate more moves per item than its *LR* counterpart, as shown in Figure 13.

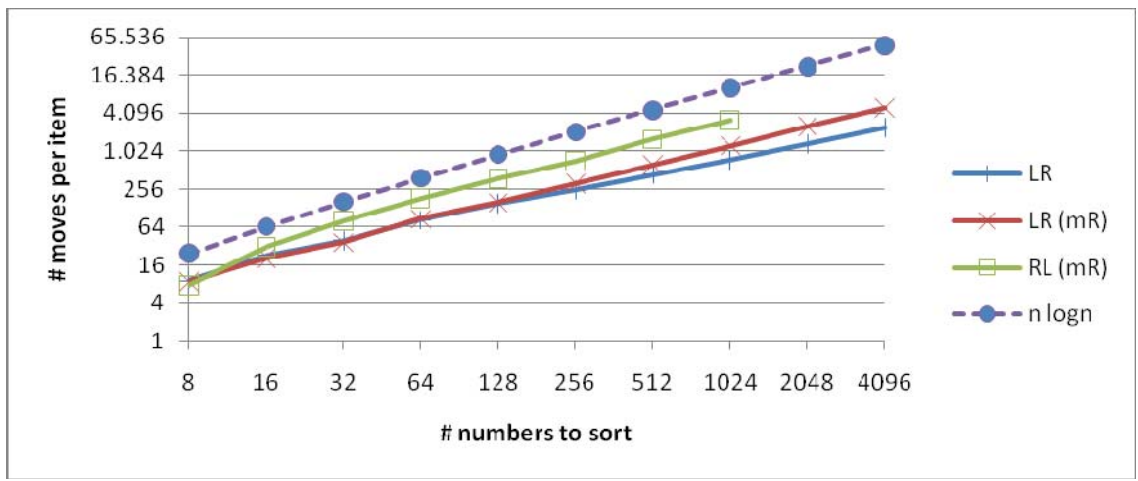

Figure 13: Constrained triple momentum must be compatible with sequence momentum.

## 4 Experimental Validation of Emerge-Sort in Almost Pre-sorted Sequences

Emerge-Sort is supposed to be able to sense the emergent direction for sorting an un-ordered sequence and then direct big and small values towards opposing ends of that sequence.

In this section we examine a smaller problem, yet very interesting: given a sorted sequence, which we then tweak to remove the ordering, how easy (or difficult) is it for Emerge-Sort to alleviate these tweaks and result in a sorted sequence?

We use the *LR* baseline to compare our results to, and it is only *LR* variants that we are now interested in. In all experiments we start with a sorted sequence and experiment with two broad families of order-tweaking operators:

- Shuffle (*k*): *k* sequence elements are selected at random and swapped with their neighbour
- Reverse (*k*): a random subsequence of length *k* is reversed

We first review the shuffling case. Figure 14 suggests that results are relatively irrelevant of the number of shuffles performed, as long as these are more than a "few". These results are also confirmed by Figure 15, where we observe again that the overhead gap is not linear, since a "few" shuffles (log*n*, in our case) can be contained with relatively constant overhead. It is an interesting question at which shuffling percentage one shifts to paying the full price.

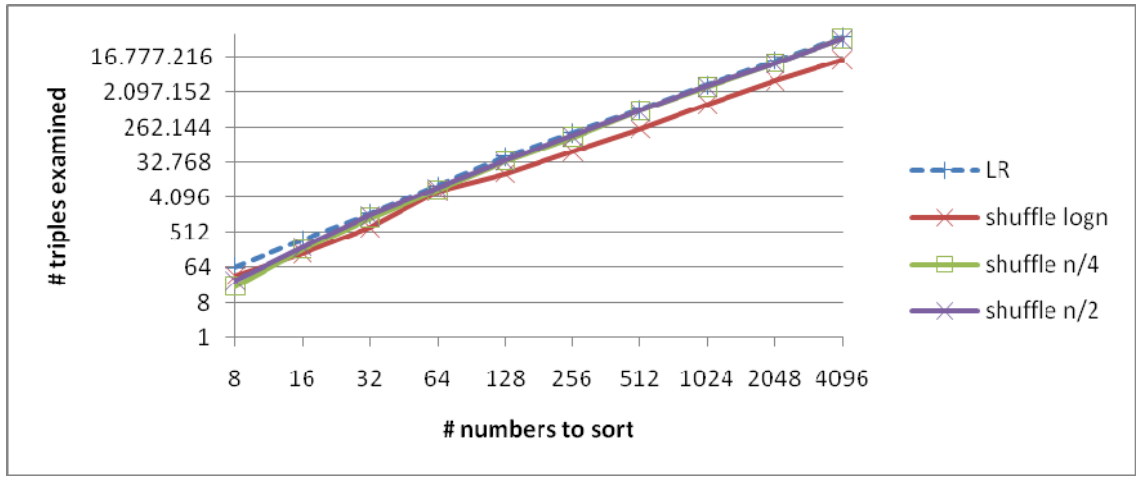

Figure 14: How shuffling affects the number of triples examined.

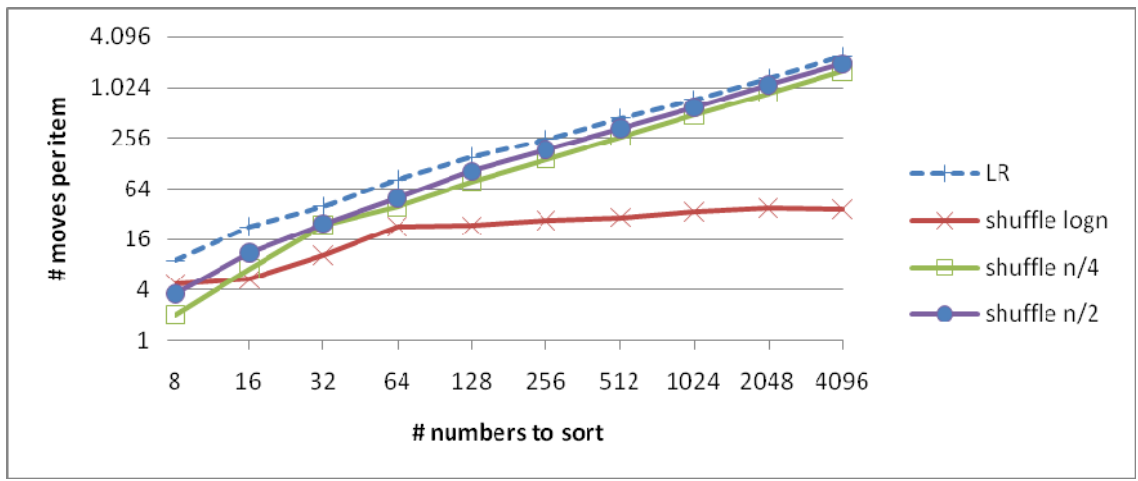

Figure 15: How shuffling affects the number of moves performed.

The reversing case is also very interesting. Figure 16 suggests that results are proportional to the length of the reversed sequence (which is not surprising) and that, besides reversing a subsequence of just a "few" numbers, these proportion observation carries on to the average number of moves per item (note the scaled values in Y-axis to avoid showing the logarithms of numbers near zero).

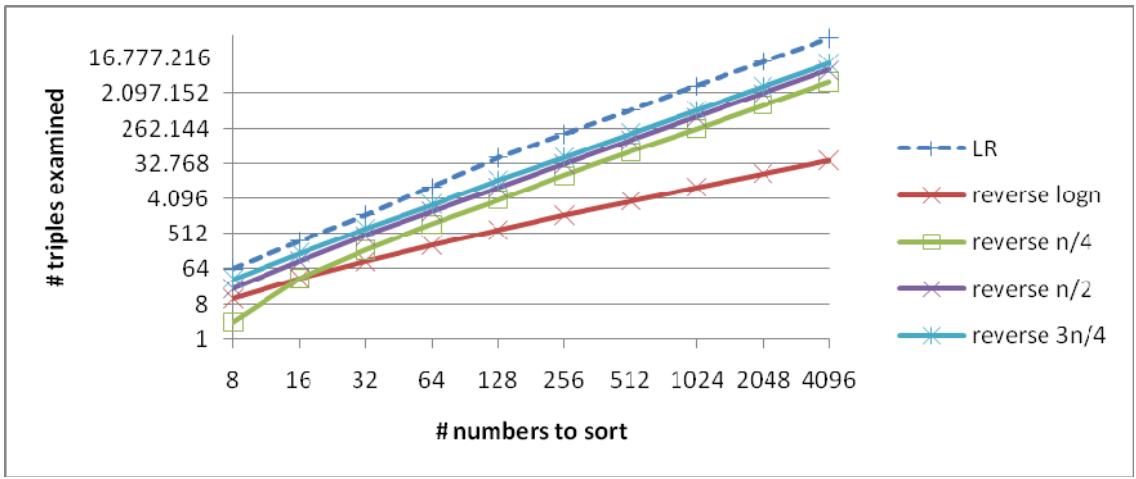

Figure 16: How reversing affects the number of triples examined.

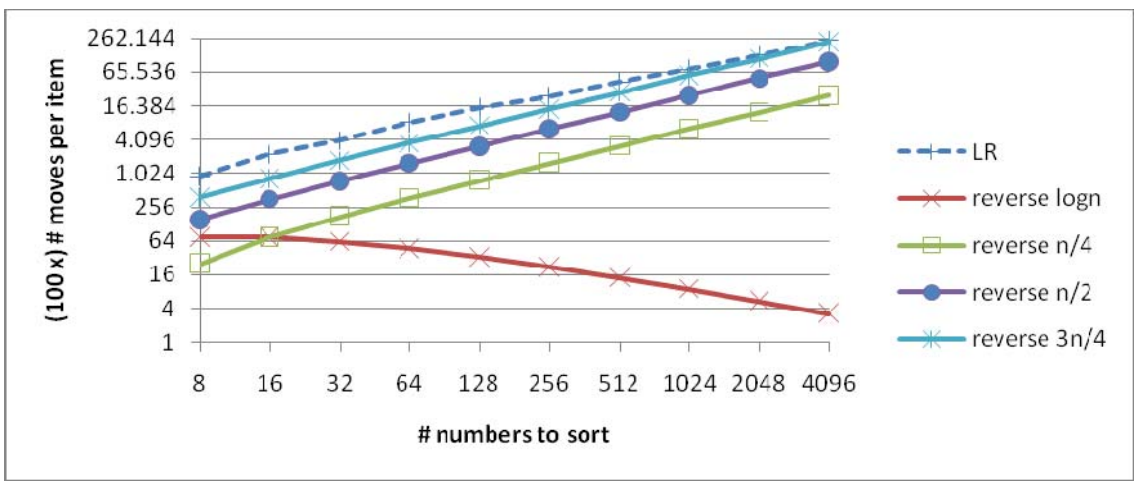

Figure 17: How reversing affects the number of moves performed.

There was one counter-intuitive observation which triggered further experimentation: reversing a sequence of 3$n$/4 elements, we initially expected, would result in expending resources comparable to what was observed when reversing a sequence of $n$/4 elements, since in both cases 75% of the sample would be sorted. The rationale, we thought, was that both operations seem to create the same level of disorder in the original sorted sequence. When this was not confirmed we realised that the reason for that behaviour was that reversing usually happens at the middle of the sequence. In those cases the largest-value and smallest-value anchors do not change and, hence, we indeed end up in a sequence that cannot reverse its sorting order yet has many elements in an order that must be totally reversed.

## 5  A Brief Discussion on Validity and Pending questions

Emerge-sort is not yet an algorithm: while we have shown promising results as far as experimentation goes, we have not yet proved properties that will elevate it to algorithm status. That particular question may have to be answered in parts, but the central question is whether we can prove that a selection of the local operators (plus or minus some variations or features) will eventually result in a sorted sequence. We have reported on how some operators seem to affect whether we will get a result or not or whether we will get a result fast or slowly, but we have not investigated the extent to which some variations of the local operators might deliver a firm theoretical result.

Settling the issue of the algorithmic nature will also give rise to the problem of analysing the complexity of Emerge-sort. Therein we must also address the issue of whether we can relate the complexity of Emerge-sort to a measurable disorder of the examined sequences (Manilla, 1985; Estivill-Castro and Wood, 1992; Biedl *et al.*, 2004).

The latter relation suggests that examination is also warranted on the mechanism by which we detect whether a sequence is sorted. Right now, we have focused on implementing the local operators that induce the sorting and have implemented a simple one-pass algorithm to detect order. This pass is performed after each round of local operators and is not implemented in the local operators. In other words, while a local operator may eternally instruct a number triple to remain intact, it has no way of telling that triple that this is also a global feature of the sequence and that it will not have to revert itself in the future. This has the interesting side-effect that if the sequence is changed, then an element of autonomic computing emerges and the sequence resorts itself. Nevertheless, we need to investigate this mechanism in depth.

A further issue that may also be interesting (initially, from an experimental point of view) is whether the the $n/\log n$ delay coefficient is related to the size of the local operators (in our case, we have triples that may affect also affect their neighbours). If we extend the scope of a local operator to a quadruple, it is reasonable to ask whether we inject unnecessarily much local overhead compared to what we might improve in the price of anarchy we pay. Alternatively, we may ask whether $n/\log n$ is a delay coefficient that characterizes all local operators. In essence, this question raises the issue of trading off computational efficiency for cost of long-range connections, as is described by Mitchell (2006) and shown to be an area of significant research promises.

## 6 Conclusions and future directions

Emerge-sort is the result of an experimental effort into investigating the dynamics of local operators regarding their ability to generate order. As such it now has more of an investigative interest, especially so since one can observe it bearing so many relations to other disciplines in computing. We expect that by studying it alongside established computing paradigms, we will gain more insight into what constitute the crossing lines between such paradigms.

## Acknowledgements

The Maple code of Emerge-Sort and all results are available on demand for academic purposes.